\documentclass{article}
\usepackage{spconf,amsmath,amssymb,graphicx}
\usepackage{multirow}
\providecommand{\reals}	{\mathbb{R}}
\ninept

\newcommand{\argmin}{\operatornamewithlimits{argmin}}

\providecommand{\myboldg}[1]    {\mbox{\boldmath $#1$}}         
\providecommand{\mybolda}       {\myboldg}                      
\providecommand{\myboldA}       {\mybolda}                      
\providecommand{\myboldn}       {\mybolda}                      

\providecommand{\boldscript}[1] {\mbox{\boldmath\scriptsize$#1$}}

\providecommand{\Bone}  {\myboldn{1}}

\providecommand{\Bh}{\mybolda{h}}

\providecommand{\Bq}{\mybolda{q}}

\providecommand{\Bs}{\mybolda{s}}

\providecommand{\Bw}{\mybolda{w}}
\providecommand{\Bx}{\mybolda{x}}
\providecommand{\By}{\mybolda{y}}

\providecommand{\bB}{\myboldA{B}}

\providecommand{\bG}{\myboldA{G}}

\providecommand{\bS}{\myboldA{S}}

\providecommand{\bV}{\myboldA{V}}
\providecommand{\bW}{\myboldA{W}}

\providecommand{\bY}{\myboldA{Y}}

\providecommand{\Btheta}        {\myboldg{\theta}}

\title{DEEP NEURAL NETWORKS FOR SINGLE CHANNEL SOURCE SEPARATION}
\name{Emad M. Grais, Mehmet Umut Sen, Hakan Erdogan}
\address{Faculty of Engineering and Natural Sciences,\\
Sabanci University, Orhanli Tuzla, 34956, Istanbul.\\
\{grais, umutsen, haerdogan\}@sabanciuniv.edu}
\begin{document}
%
\maketitle
\begin{abstract}
In this paper, a novel approach for single channel source separation (SCSS) using a deep neural network (DNN) architecture is introduced. Unlike previous studies in which DNN and other classifiers were used for classifying time-frequency bins to obtain hard masks for each source, we use the DNN to classify estimated source spectra to check for their validity during separation. In the training stage, the training data for the source signals are used to train a DNN. In the separation stage, the trained DNN is utilized to aid in estimation of each source in the mixed signal. Single channel source separation problem is formulated as an energy minimization problem where each source spectra estimate is encouraged to fit the trained DNN model and the mixed signal spectrum is encouraged to be written as a weighted sum of the estimated source spectra. The proposed approach works regardless of the energy scale differences between the source signals in the training and separation stages. Nonnegative matrix factorization (NMF) is used to initialize the DNN estimate for each source. The experimental results show that using DNN initialized by NMF for source separation improves the quality of the separated signal compared with using NMF for source separation.  

\end{abstract}
\section{Introduction}
\label{sec:intro}

Single channel audio source separation is an important and challenging problem and has received considerable interest in the research community in recent years. Since there is limited information in the mixed signal, usually one needs to use training data for each source to model each source and to improve the quality of separation. In this work, we introduce a new method for improved source separation using nonlinear models of sources trained using a deep neural network.

\subsection{Related work}

Many approaches have been introduced so far to solve the single channel source separation problem. Most of those approaches strongly depend on training data for the source signals. The training data can be modeled using probabilistic models such as Gaussian mixture model (GMM) \cite{Kristjansson:04:smssuhrsr,Reddy:07:smm, Reddy:04:amseefscss}, hidden Markov model (HMM) or factorial HMM \cite{Virtanen:06:srufhmmfsitfs,Roweis:00:omss,Deoras:04:afhatsroidsbmtooac}. These models are learned from the training data and usually used in source separation under the assumption that the sources appear in the mixed signal with the same energy level as they appear in the training data. Fixing this limitation requires complicated computations as in  \cite{radfar:09:geimbscss,radfar:08:mssbogammsee,radfar:07:ltgeimbscss, ozerov:09:fshmmfparass, radfar:10:sfhmm, hershey:10:shm}. Another approach to model the training data is to train nonnegative dictionaries for the source signals \cite{Schmidt:06:scssusnnmf,emad:12:stpsnmfscss,emad:11:scsmsunmfwswasm}. This approach is more flexible with no limitation related to the energy differences between the source signals in training and separation stages. The main problem in this approach is that any nonnegative linear combination of the trained dictionary vectors is a valid estimate for a source signal which may decrease the quality of separation. Modeling the training data with both nonnegative dictionary and cluster models like GMM and HMM was introduced in \cite{ grais:12:rnmfgmpsscss, emad:12:gmgprnmfscss, emad:12:hmmprnmfscss, emad:13:stpemenbscss} to fix the limitation related to the energy scaling between the source signals and training more powerful models that can fit the data properly. Another type of approach which is called classification-based speech separation aims to find hard masks where each time-frequency bin is classified as belonging to either of the sources. For example in \cite{ Yoshii:12:cppvsp}, various classifiers based on GMM, support vector machines, conditional random fields, and  deep neural networks were used for classification.

\subsection{Contributions}

In this paper, we model the training data for the source signals using a single joint deep neural network (DNN). The DNN is used as a spectral domain classifier which can classify its input spectra into each possible source type. Unlike classification-based speech separation where the classifiers are used to segment time-frequency bins into sources, we can obtain soft masks using our approach. Single channel source separation problem is formulated as an energy minimization problem where each source spectral estimate is encouraged to fit the trained DNN model and the mixed signal spectrum is encouraged to be  written as a weighted sum of the estimated source spectra. Basically, we can think of the DNN as checking whether the estimated source signals are lying in their corresponding nonlinear manifolds which are represented by the trained joint DNN. Using a DNN for modeling the sources and handling the energy differences in training and testing is considered to be the main novelty of this paper. Deep neural network (DNN) is a well known model for representing the detailed structures in complex real-world data \cite{hinton:06:afl, hinton:12:dnn}. 
Another novelty of this paper is using nonnegative matrix factorization \cite{emad:11:scsmsunmfasm} to find initial estimates for the sources rather than using random initialization. 

\subsection{Organization of the paper}
This paper is organized as follows: In Section \ref{pf} a mathematical formulation for the SCSS problem is given. Section \ref{sec:nmf} briefly describes the NMF method for source separation. In Section \ref{sec:meth}, we introduce our new method. We present our experimental results in Section \ref{sec:exp}. We conclude the paper in Section \ref{sec:conc}.

\section{Problem formulation}
\label{pf}
In single channel source separation problems, the aim is to find estimates of source signals that are mixed on a single channel $y(t)$. For simplicity, in this paper we assume the number of sources is two. This problem is usually solved in the short time Fourier transform (STFT) domain. Let $Y(t,f)$ be the STFT of $y(t)$, where $t$ represents the frame index and $f$ is the frequency-index. Due to the linearity of the STFT, we have:   
\begin{equation}
\label{stftsum}
Y(t,f)=S_1(t,f)+S_2(t,f),
\end{equation}
where $S_1(t,f)$ and $S_2(t,f)$ are the unknown STFT of the first and second sources in the mixed signal. In this framework~\cite{Schmidt:06:scssusnnmf,Virtanen:07:msssbnmfwtcasc}, the phase angles of the STFT were usually ignored. Hence, we can approximate the magnitude spectrum of the measured signal as the sum of source signals' magnitude spectra as follows:
\begin{equation}
\label{stftangl_2}
\left|Y(t,f)\right|\approx\left| S_1(t,f)\right|+\left| S_2(t,f)\right|.
\end{equation}
We can write the magnitude spectrogram in matrix form as follows:
\begin{equation}
\bY \approx \bS_1+\bS_2.
\end{equation} 
where $\bS_1,\bS_2$ are the unknown magnitude spectrograms of the source signals and need to be estimated using the observed mixed signal and the training data.

\section{NMF for supervised source separation}
\label{sec:nmf}
In this section, we briefly describe the use of nonnegative matrix factorization (NMF) for supervised single channel source separation. We will relate our model to the NMF idea and we will use the source estimates obtained from using NMF as initilization for our method, so it is appropriate to introduce the use of NMF for source separation first.

To find a suitable initialization for the sources signals, we use nonnegative matrix factorization (NMF) as in \cite{emad:11:scsmsunmfasm}. NMF \cite{Lee:01:afnmf} factorizes any nonnegative matrix $\bV$ into a basis matrix (dictionary) $\bB$ and a gain matrix $\bG$ as
\begin{equation}
\label{NMF_mat}
	\bV\approx\bB\bG.
\end{equation}
The matrix $\bB$ contains the basis vectors that are optimized to allow the data in $\bV$ to be approximated as a linear combination of its constituent columns. The solution for $\bB$ and $\bG$ can be found by minimizing the following Itakura-Saito (IS) divergence cost function~\cite{Fevotte:09:nmfwisd}:
\begin{equation}
\label{div}
\footnotesize{
\min_{\bB,\bG} D_{IS}\left(\bV\left|\right|\bB\bG\right),	
}
\end{equation}
where 
\[
\footnotesize{
D_{IS}\left(\bV\left|\right|\bB\bG\right)=\sum_{a,b}\left(\frac{\bV_{a,b}}{\left(\bB\bG\right)_{a,b}}-\log\frac{\bV_{a,b}}{\left(\bB\bG\right)_{a,b}}- 1\right).
}
\] 
This divergence cost function is a good measurement for the perceptual difference between different audio signals \cite{Fevotte:09:nmfwisd}. The IS-NMF solution for equation~(\ref{div}) can be computed by alternating multiplicative updates of $\bG$ and $\bB$ as follows:
\begin{equation}
\label{weights}
\footnotesize
{
\bG\leftarrow \bG \otimes \frac{\bB^T\left(\frac{\bV}{\left(\bB\bG\right)^2}\right)}{\bB^T\left(\frac{\Bone}{\bB\bG}\right)},	
}
\end{equation}
\begin{equation}
\label{basis}
\footnotesize
{
\bB\leftarrow \bB \otimes \frac{\left(\frac{\bV}{\left(\bB\bG\right)^2}\right)\bG^T}{\left(\frac{\Bone}{\bB\bG}\right)\bG^T},	
}
\end{equation} 
where $\Bone$ is a matrix of ones with the same size of $\bV$, the operation $\otimes$ is an element-wise multiplication, all divisions and $\left(.\right)^2$ are element-wise operations. The matrices $\bB$ and $\bG$ are usually initialized by positive random numbers and then updated iteratively using equations (\ref{weights}) and (\ref{basis}).

In the initialization stage, NMF is used to decompose the frames for each source $i$ into a multiplication of a nonnegative dictionary $\bB_i$ and a gain matrix $\bG_i $ as follows:
\begin{equation}
\label{train_NMF}
	\bS_i^{train}\approx\bB_i\bG_i^{train}, \ \ \ \forall i\in\left\{1,2\right\},
\end{equation}
where $\bS_i^{train}$ is the nonnegative matrix that contains the spectrogram frames of the training data of source $i$.
After observing the mixed signal, we calculate its spectrogram $\bY_{psd}$.  NMF is used to decompose the mixed signal's spectrogram matrix $\bY_{psd}$ with the trained dictionaries as follows:
\begin{equation}
\label{decomp}
\bY_{psd}\approx\left[\bB_1,\bB_2\right]\bG\ \ \ \mbox{or} \ \ \ 
\bY_{psd}\approx
\left[
\bB_1 \ \ \ \bB_2
\right]\left[
\begin{array}[pos]{c}
\bG_1 \\
\bG_2
\end{array}	
\right].
\end{equation}
The only unknown here is the gains matrix $\bG$ since the dictionaries are fixed. The update rule in equation (\ref{weights}) is used to find $\bG$. After finding the value of $\bG$, the initial estimate for each source magnitude spectrogram is computed as follows:
\begin{equation}
\label{sest1}
\footnotesize
{
\hat {\bS}_{{init}_1}=\frac{\bB_1\bG_1}{\bB_1\bG_1+\bB_2\bG_2}\otimes\bY, \ \ \ \hat {\bS}_{{init}_2}=\frac{\bB_2\bG_2}{\bB_1\bG_1+\bB_2\bG_2}\otimes\bY,
}
\end{equation}
where $\otimes$ is an element-wise multiplication and the divisions are done element-wise.
 
The magnitude spectrograms of the initial estimates of the source signals are used to initialize the sources in the separation stage of the DNN approach.


\section{The method}
\label{sec:meth}
In NMF, the basic idea is to model each source with a dictionary, so that source signals appear in the nonnegative span of this dictionary. In the separation stage, the mixed signal is expressed as a nonnegative linear combination of the source dictionaries and separation is performed by taking the parts corresponding to each source in the decomposition.

The basic problem in NMF is that each source is modeled to lie in a nonnegative cone defined by all the nonnegative {\em linear} combinations of its dictionary entries. This assumption may be a limiting assumption usually since the variability within each source indicates that nonlinear models may be more appropriate. This limitation led us to consider nonlinear models for each source. It is not trivial to use nonlinear models or classifiers in source separation. 
Since deep neural networks were recently used with increased success in speech recognition and other object recognition tasks, they can be considered as superior models of highly variable real-world signals. 

We first train a DNN to model each source in the training stage. We then use an energy minimization objective to estimate the sources and their gains during the separation stage. Each stage is explained below.

\subsection{Training the DNN}
\begin{figure}[h]
\begin{center}
\includegraphics[width=0.3\linewidth]{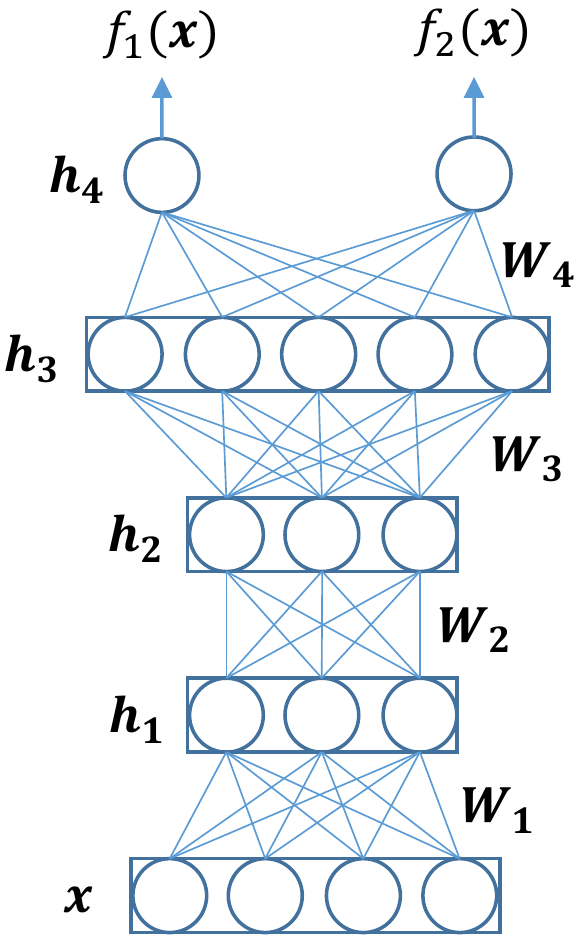}
\caption{\label{fig:DNN2} Illustration of the DNN architecture.}
\end{center}
\end{figure}

We train a DNN that can classify sources present in the mixed signal. The input to the network is a frame of normalized magnitude spectrum, $\Bx \in \reals^d$. The neural network architecture is illustrated in Figure \ref{fig:DNN2}. There are two outputs in the DNN, each corresponding to a source. The label of each training instance is a binary indicator function, namely if the instance is from source one, the first output label $f_1(\Bx)=1$ and the second output label $f_2(\Bx) = 0$. 
Let $n_k$ be the number of hidden nodes in layer $k$ for $k=0,\ldots,K$ where $K$ is the number of layers. Note that $n_0=d$ and $n_K=2$. Let $\bW_k \in \reals^{n_{k}\times n_{k-1}}$ be the weights between layers $k-1$ and $k$, then the values of a hidden layer $\Bh_k \in \reals^{n_k}$ are obtained as follows:
 \begin{equation}
\Bh_k = g(\bW_k\Bh_{k-1}),
\end{equation}
where $g(x) = \frac{1}{1+\exp(-x)}$ is the elementwise sigmoid function. We skip the bias terms to avoid clutter in our notation. The input to the network is $h_0 = \Bx \in \reals^d$ and the output is $f(\Bx) = \Bh_K \in \reals^2$.

Training a deep network necessitates a good initialization of the parameters. It is shown that layer-by-layer pretraining using unsupervised methods for initialization of the parameters results in superior performance as compared to using random initial values. We used Restricted Boltzmann Machines (RBM) for initialization. After initialization, supervised backpropagation algorithm is applied to fine-tune the parameters. The learning criteria we use is least-squares minimization.  We are able to get the partial derivatives with respect to the inputs, and this derivative is also used in the source separation part. Let $f(.):\reals^d \rightarrow \reals^2$ be the DNN, then $f_1(\Bx)$ and $f_2(\Bx)$ are the scores that are proportional to the probabilities of source one and source two respectively for a given frame of normalized magnitude spectrum $\Bx$. We use these functions to measure how much the separated spectra carry the characteristics of each source as we elaborate more in the next section.

\subsection{Source separation using DNN and energy minimization}
In the separation stage, our algorithm works independently in each frame of the mixed audio signal.
For each frame of the mixed signal spectrum, we calculate the normalized magnitude spectrum $\By$. We would like to express $\By=u\Bx_1+v\Bx_2$ where $u$ and $v$ are the gains and $\Bx_1$ and $\Bx_2$ are normalized magnitude spectra of source one and two respectively.

We formulate the problem of finding the unknown parameters $\Btheta=(\Bx_1,\Bx_2,u,v)$ as an energy minimization problem. We have a few different criteria that the source estimates need to satisfy. First, they must fit well to the DNN trained in the training stage. Second, their linear combination must sum to the mixed spectrum $\By$ and third the source estimates must be nonnegative since they correspond to the magnitude spectra of each source.

The energy functions $E_1$ and $E_2$ below are least squares cost functions that quantify the fitness of a source estimate $\Bx$ to each corresponding source model in the DNN.
\begin{equation}
E_1(\Bx) = (1-f_1(\Bx))^2 + (f_2(\Bx))^2,
\end{equation}

\begin{equation}
E_2(\Bx) = (f_1(\Bx))^2 + (1-f_2(\Bx))^2.
\end{equation}
Basically, we expect to have $E_1(\Bx) \approx 0$ when $\Bx$ comes from source one and vice versa.
We also define the following energy function which quantifies the energy of error caused by the least squares difference between the mixed spectrum $\By$ and its estimate found by linear combination of the two source estimates $\Bx_1$ and $\Bx_2$:
\begin{equation}
E_{err}(\Bx_1,\Bx_2,\By,u,v) = ||u\Bx_1+v\Bx_2-\By||^2.
\end{equation}
Finally, we define an energy function that measures the negative energy of a variable,
$
R(\theta) = (\min(\theta,0))^2.
$

In order to estimate the unknowns in the model, we solve the following energy minimization problem.
\begin{equation}
(\hat{\Bx}_1,\hat{\Bx}_2,\hat u,\hat v) = \argmin_{\{\boldscript{\Bx_1},\boldscript{\Bx_2},u,v\}} E(\Bx_1,\Bx_2,\By,u,v),
\end{equation}
where
\begin{equation}
\begin{split}
E(\Bx_1,\Bx_2,\By,u,v) &= E_1(\Bx_1) + E_2(\Bx_2) + \lambda  E_{err}(\Bx_1,\Bx_2,\By,u,v)  \\
           &+ \beta\sum_{i}{R(\theta_i)}
\end{split}
\end{equation}
is the joint energy function which we seek to minimize. $\lambda$ and $\beta$ are regularization parameters which are chosen experimentally. Here 
$\Btheta=(\Bx_1,\Bx_2,u,v)=[\theta_1,\theta_2,\ldots,\theta_n]$ is a vector containing all the unknowns which must all be nonnegative. Note that, the nonnegativity can be given as an optimization constraint as well, however we obtained faster solution of the optimization problem if we used the negative energy function penalty instead. If some of the parameters are found to be negative after the solution of the optimization problem (which rarely happens), we set them to zero. We used the LBFGS algorithm for solving the unconstrained optimization problem.

We need to calculate the gradient of the DNN outputs with respect to the input $\Bx$ to be able to solve the optimization problem. 
The gradient of the input $\Bx$ with respect to $f_i(\Bx)$ is given as $\frac{\partial f_i(\Bx)}{\partial \Bx}=\Bq_{1,i}$ for $i=1,2$, where,
 
\begin{equation}
\Bq_{k,i} = \bW_k^T(\Bq_{k+1,i} \otimes \Bh_k \otimes (1-\Bh_k)),
\end{equation}
and $\Bq_{K,i} = f_i(\Bx)(1-f_i(\Bx))\Bw_{K,i}^T$, where $\Bw_{K,i} \in \reals^{n_{K-1}}$ contains the weights between $i^{th}$ node of the output layer and the nodes at the previous layer, in other words the $i$th row of $\bW_K$. 

\begin{figure}[h]
\begin{center}
\includegraphics[width=1\linewidth]{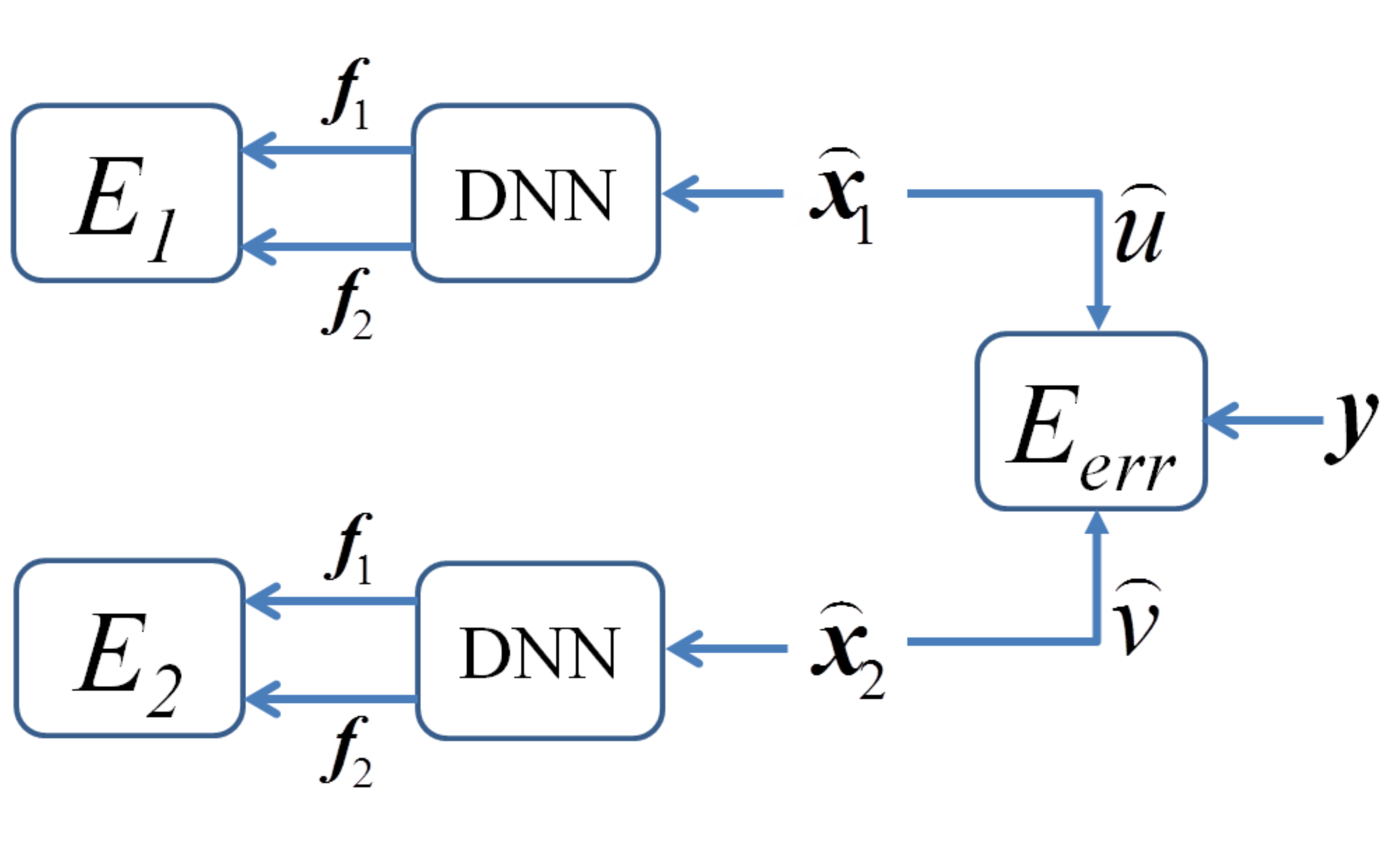}
\caption{\label{fig:chart}Flowchart of the energy minimization setup. For illustration, we show the single DNN in two separate blocks in the flowchart.}
\end{center}
\end{figure}

The flowchart of the energy minimization setup is shown in Figure \ref{fig:chart}. For illustration, we show the single DNN in two separate blocks in the flowchart. The fitness energies are measured using the DNN and the error energy is found from the summation requirement.

Note that, since there are many parameters to be estimated and the problem is clearly non-convex, the initialization of the parameters is very important. We initialize the estimates $\hat{\Bx}_1$ and $\hat{\Bx}_2$ from the NMF result after normalizing by their $\ell_2$-norms. $\hat u$ is initialized by the $\ell_2$-norm of the initial NMF source estimate $\hat{\Bs}_1$ divided by the $\ell_2$-norm of the mixed signal $\By$. $\hat v$ is initialized in a similar manner.

After we obtain $(\hat{\Bx}_1,\hat{\Bx}_2,\hat u,\hat v)$ as the result of the energy minimization problem, we use them as spectral estimates in a Wiener filter to reconstruct improved estimates of each source spectra, {\em e.g.} for source one we obtain the final estimate as follows:

\begin{equation}
\label{sest1}
\hat {\Bs}_{1}=\frac{(\hat u \hat\Bx_1)^2}{(\hat u \hat\Bx_1)^2+(\hat v \hat \Bx_2)^2}\otimes\By.
\end{equation}

\section{Experiments and Discussion}
\label{sec:exp}
We applied the proposed algorithm to separate speech and music signals from their mixture. We simulated our algorithm on a collection of speech and piano data at 16kHz sampling rate. For speech data, we used the training and testing male speech data from the TIMIT database. For music data, we downloaded piano music data from piano society web site \cite{site:pianosociety}. We used 39 pieces with approximate 185 minutes total duration from different composers but from a single artist for training and left out one piece for testing. The magnitude spectrograms for the speech and music data were calculated using the STFT: A Hamming window with 480 points length and $60\%$ overlap was used and the FFT was taken at 512 points, the first 257 FFT points only were used since the conjugate of the remaining 255 points are involved in the first points.

The mixed data was formed by adding random portions of the test music file to 20 speech files from the test data of the TIMIT database at different speech to music ratio. The audio power levels of each file were found using the ``speech voltmeter'' program from the G.191 ITU-T STL software suite \cite{site:itustl}. 

For the initialization of the source signals using nonnegative matrix factorization, we used a dictionary size of 128 for each source. For training the NMF dictionaries, we used 50 minutes of data for music and 30 minutes of the training data for speech. For training the DNN, we used a total 50 minute subset of music and speech training data for computational reasons. 

For the DNN, the number of nodes in each hidden layer were 100-50-200 with three hidden layers. Sigmoid nonlinearity was used at each node including the output nodes. DNN was initialized with RBM training using contrastive divergence. We used 150 epochs for training each layer's RBM. We used 500 epochs for backpropagation training. The first five epochs were used to optimize the output layer keeping the lower layer weights untouched. 

In the energy minimization problem, the values for the regularization parameters were $\lambda=5$ and $\beta=3$. We used Mark Schmidt's minFunc matlab LBFGS solver for solving the optimization problem \cite{site:minfun}.

Performance measurements of the separation algorithm were done using the signal to distortion ratio (SDR) and the signal to interference ratio (SIR) \cite{vincent:06:pmi}. The average SDR and SIR over the 20 test utterances are reported. The source to distortion ratio (SDR) is defined as the ratio of the target energy to all errors in the reconstructed signal. The target signal is defined as the projection of the predicted signal onto the original speech signal. Signal to interference ratio (SIR) is defined as the ratio of the target energy to the interference error due to the music signal only. The higher SDR and SIR we measure the better performance we achieve. We also use the output SNR as additional performance criteria.

The results are presented in Tables \ref{table:res} and \ref{table:res2}. We experimented with multi-frame DNN where the inputs to the DNN were taken from $L$ neighbor spectral frames for both training and testing instead of using a single spectral frame similar to \cite{emad:11:scsmsunmfwswasm}. We can see that using the DNN and the energy minimization idea, we can improve the source separation performance for all input speech-to-music ratio (SMR) values from -5 to 5 dB. In all cases, DNN is better than regular NMF and the improvement in SDR and SNR is usually around 1-1.5 dB. However, the improvement in SIR can be as high as 3 dB which indicates the fact that the introduced method can decrease remaining music portions in the reconstructed speech signal. We performed experiments with $L=3$ neighboring frames which improved the results as compared to using a single frame input to the DNN. For $L=3$, we used 500 nodes in the third layer of the DNN instead of 200. We conjecture that better results can be obtained if higher number of neighboring frames are used.

\begin{table}[ht]
\caption{SDR, SIR and SNR in dB for the estimated speech signal.}
\centering
\scalebox{0.8}
{
\begin{tabular}{||c||c|c|c||c|c|c||c|c|c||}
\hline\hline
SMR & \multicolumn{3}{|c||}{NMF} & \multicolumn{6}{|c||}{DNN} \\
    & \multicolumn{3}{|c||}{ } & \multicolumn{3}{|c||}{$L=1$} & \multicolumn{3}{|c||}{$L=3$}\\
dB  & SDR&SIR &SNR& SDR&SIR&SNR& SDR&SIR&SNR \\        
\hline
-5   & 1.79  & 5.01  &  3.15    & 2.81    &  7.03 & 3.96 & {\bf3.09 } & {\bf7.40}  & {\bf4.28}\\
\hline
0    & 4.51  & 8.41  &  5.52   &  5.46  & 9.92 & 6.24 & {\bf 5.73} & {\bf 10.16}  & {\bf 6.52}\\
\hline
5    & 7.99  & 12.36  &  8.62  & 8.74  & {\bf13.39} & 9.24   & {\bf 8.96} &   13.33 & {\bf 9.45}\\
\hline  
\hline
\end{tabular}
}
\label{table:res} 
\end{table}

\begin{table}[ht]
\caption{SDR, SIR and SNR in dB for the estimated music signal.}
\centering
\scalebox{0.8}
{
\begin{tabular}{||c||c|c|c||c|c|c||c|c|c||}
\hline\hline
SMR & \multicolumn{3}{|c||}{NMF} & \multicolumn{6}{|c||}{DNN} \\
    & \multicolumn{3}{|c||}{ } & \multicolumn{3}{|c||}{$L=1$} & \multicolumn{3}{|c||}{$L=3$}\\
dB  & SDR&SIR &SNR& SDR&SIR&SNR& SDR&SIR&SNR \\        
\hline
-5   & 5.52  & 15.75  &  6.30    & 6.31    &  {\bf18.48} & 7.11 & {\bf 6.67} & 18.30  & {\bf 7.43}\\
\hline
0    & 3.51  & 12.65  &  4.88   &  4.23  & {\bf16.03} & 5.60 & {\bf 4.45} &  15.90  & {\bf 5.88}\\
\hline
5    & 0.93  & 9.03  &  3.35  & 1.79  & 12.94 & 3.96 & {\bf 1.97} &  {\bf 13.09} & {\bf 4.17}\\
\hline  
\hline
\end{tabular}
}
\label{table:res2} 
\end{table} 
%
%
%
%

\section{Conclusion}
\label{sec:conc}
In this work, we introduced a novel approach for single channel source separation (SCSS) using deep neural networks (DNN). The DNN was used in this paper as a helper to model each source signal. The training data for the source signals were used to train a DNN. The trained DNN was used in an energy minimization framework to separate the mixed signals while also estimating the scale for each source in the mixed signal. Many adjustments for the model parameters can be done to improve the proposed SCSS using the introduced approach. Different types of DNN such as deep autoencoders and deep recurrent neural networks which can handle the temporal structure of the source signals can be tested on the SCSS problem. We believe this idea is a novel idea and many improvements will be possible in the near future to improve its performance.


\vfill\pagebreak


\bibliographystyle{IEEEbib}
\bibliography{strings,refs}

\end{document}